%% file: 0-main-data-foundations.tex
\documentclass[runningheads]{llncs}
\usepackage[T1]{fontenc}
\usepackage[utf8]{inputenc}
\usepackage{times}
\usepackage{hyperref}
\usepackage{amsmath}
\usepackage{amssymb}
\usepackage{subcaption}
\usepackage{graphicx}
\usepackage{algorithm}
\usepackage{algpseudocode}
\usepackage{booktabs} 
\usepackage{enumitem}
\usepackage{todonotes}
\usepackage{threeparttable}
\usepackage{multirow}
\usepackage{pgf}
\usepackage{pgfplots}

\usepackage{xspace}

\newcommand{\eg}{\textit{e}.\textit{g}.,\xspace}

\begin{document}


%
\title{Reviving our data foundations is the most disruptive step to data maturity}
\titlerunning{Reviving our data foundations is the most disruptive step to data maturity}
%
\author{
Valentina Carapella\thanks{Corresponding author.} 
\and
Ernesto Jiménez-Ruiz
}
%
%
\institute{
City St George’s, University of London, London, UK\\
\email{valentina.carapella@datafirstbiz.co.uk, ernesto.jimenez-ruiz@citystgeorges.ac.uk}
}

\maketitle              
\begin{abstract}

The most disruptive step that enterprises of small-medium size and maturity can take to make the most of the latest technological advances in AI is to step back from the hype and focus on establishing or reviving a good knowledge foundation layer. It is a hard message to present to the executive team; therefore, it needs to be backed by evidence, and its implementation needs to be of minimal impact on the existing processes. In this vision statement, we discuss how we need to rethink what evidence speaks to the decision-makers and propose a low-impact data strategy that adapts to the existing and ever-changing data flows and processes across the company. We firmly believe that knowledge graph techniques will increasingly become non-negotiable in the data strategy of an AI-powered enterprise, provided that we approach their design in a modular, dynamic and cross-functional way.

\medskip






\keywords{Data maturity \and Ontology \and Enterprise \and Business processes}

\end{abstract}

\setcounter{footnote}{0} 

\input{vc-intro}

\input{ejr-onto}

\section{Conclusions}
\label{sec:end}

The path to a successful adoption of AI-powered solutions for start-up to scale-up companies is paved by their data maturity. Knowledge graph technologies are a key technical enabler but investing in major data management projects is a hard ask on most companies with limited resources and challenging business priorities. Our vision for a successful adoption is one of minimal investment with maximal gain, focussing on light-weight, dynamic representation of processes and workflows and their modes of change, high modularity and cross-functional mapping of semantics. The best way to navigate the technology hype is to take a step back and consolidate the data foundations and its good practices. This is not a new message by all means, but one that is hard for most companies to embrace when the promise and pressure of quick gains from the latest shiny technology on the market are so high. Our most heartfelt recommendation is to refrain from the hype and focus on foundational work, this is truly the most disruptive step a company can take. 





\subsubsection*{Declaration of use of Generative AI.}
AI tools were used solely for grammar correction and minor language edits. They did not contribute to content creation, idea development, or substantive rewriting. The research design, experiments, analysis, and writing were carried out entirely by the authors.

\bibliographystyle{splncs04}
\bibliography{vc-ref}
\end{document}

%% file: vc-intro.tex
\section{Introduction}
\label{sec:context}

Having experienced multiple hype cycles in the technical industry, we are used to companies of different size and maturity rushing to embrace the latest tool or advancement, only to find later on that the rate of failure in their implementation is surprisingly high. Data-driven, and more recently, AI-driven approaches are no different, because the problem does not lie with the technology, but with the adoption strategy \cite{mckinsey2026seven}. Our claim is that the most disruptive step for a company of small-mid size eager to join the AI-driven transformation that is occurring worldwide is, counter-intuitively, a step back. By laying solid data foundations across the functional units of the business and investing in minimally invasive, incremental steps towards a greater data maturity, harnessing the benefit of AI-powered technical solutions will be achievable and sustainable.

Two aspects need to be taken into account. First, it is typically hard to gain buy-in from the executives on any kind of foundational work, aptly suggested by the most common name used to describe it, technical debt. We need to revert this view, showcasing the business value of foundational improvements. It is useful but by no means sufficient to provide evidence as to how the outcomes produced by any AI-based tool or system will only be as good as the data we feed it. This is a generally accepted truth, but is unlikely to change the Executive's opinion or steer the Product team's priorities. We need to create a compelling connection between the foundational improvements we propose and the Key Performance Indicators (KPI), such as Return On Investment (ROI), Time To Market (TTM), etc.

Academic research should not necessarily provide evidence for such an improvement in KPIs, but being aware of the parameters used to evaluate new technology will help filter among multiple promising research avenues. This in turn will open more opportunities for further funding and fruitful academia-industry collaborations.

\section{Business perspective}

In the past five years there has been a major cultural and structural shift in how data-savvy enterprises value their data and self-organise to maximise such value, formalised by Dehghani at ThoughtWorks as the Data Mesh \cite{dehghani2022data}. Based on our experience and confirmed by recent reviews \cite{bode2024toward}, the principles of the Data Mesh are the best framework proposed so far to truly unlock the value of data for a business that aims to be data-driven. We believe that knowledge graph technologies lend themselves quite naturally to enable such approach and its emphasis on domain ownership and distributed data governance. However, as mentioned in Section \ref{sec:context}, it is still difficult to gain the trust of decision makers in a good portion of small-medium-sized companies in the technical sector, due to a number of reasons, most notably a general lack of awareness, resource, and financial pressure \cite{sequeda2022designing}. 
We propose advances in three key areas that we believe will change this scenario dramatically in the next 5 years, as they help relieve the pressure on this wide portion of the technical industry by requiring minimal investment and transformation while providing maximal benefit.

\begin{enumerate}
    \item Ontologies of processes and workflows rather than just about data and relationships
    \item Modular and minimal ontologies for cross-departmental interoperability
    \item Ontology alignment to the next level, mapping cross-functionally.
\end{enumerate}
In the remainder of this Section, we will provide motivation as to why.

\subsection{Ontologies of processes and workflows rather than just about data and relationships}
\label{sec:ontoprocess}

The data life cycle is usually described by 5-8 stages covering the end-to-end flow of data across the company from its generation/ acquisition, through management and processing, to visualisation and interpretation \cite{SEBASTIANCOLEMAN2022257}. As the data travels through these stages it is consumed by actors/agents (human or systems) and transformed. Effective data governance and data lineage ensure that we can always trace back such transformations. The information that allows such tracing is captured in the metadata, which is enriched at each stage as it remains tied to the data it describes. Or at least this is the ideal scenario. The reality is more complex and often combines gaps in the information with duplicates and ambiguities. Yet, most businesses have developed through the years workarounds, ad hoc solutions and sub-processes that all contribute to the successful delivery of the final product or service. While efforts to standardise and streamline are highly recommended, in practice and in the short/mid term, they are often unfeasible. One of the reasons is that in fast-paced industry environments, such processes change continuously to accommodate for new customer requirements or business needs, and the data generated inevitably changes with them. 

The importance of the link between metadata and ontologies has been highlighted by leading data strategy consultancies \cite{carruthers2023halo}, an idea that many aspiring and established data-driven enterprises have already internalised but only in a few instances has been successfully implemented. In fast-changing environments, ontologies are only as useful as they are able to stay relevant. This means keeping a pragmatic approach and modelling the source of change and the means by which it is applied, in other words, letting the ontologies in the business describe processes and workflows, rather than focussing only on a static snapshot of data and its relationships. Metadata that is structured and produced following such ontologies will provide a more complete and reliable passport to the data as it travels through its life cycle.

\subsection{Modular and minimal ontologies for cross-departmental interoperability}
\label{sec:modular}

A common problem in fast-evolving companies, especially those that are moving from start-up to scale-up stage, is that of formalising their knowledge in a way that supports their growth. A successful investment round or product release might trigger a doubling in resource and business functions in the order of months. These are exciting times, but also dangerous ones. It is common that the need for speed forces teams to cut corners, with the most obvious candidate being documentation and foundational improvements. It is fundamental that the gathering and formalisation of knowledge is as automated and as light-weight as possible, with minimal overhead for data management. A pragmatic approach to ontology development, focusing on minimal domains and prioritising the most important business processes, is a feasible and rewarding approach. Following the domain ownership distribution across the company, and allowing for fast prototyping of ontologies, the resource overhead is minimal, the ontology provides fresh and relevant information, and the focus can be on a few key applications. This would call for a modular approach to knowledge graph generation, and a routine upgrade with changing requirements and business needs for the ontologies to stay relevant.

\subsection{Ontology alignment to the next level, mapping cross-functionally}
\label{sec:align}

Each department or functional unit in the business \textit{speaks a different language}, this might be for historical reasons but also a result of genuine business utility. For example, customer-facing departments will have a knowledge base that is heavily geared towards the needs of customers and details of products and services, whereas a knowledge base for the technical / engineering units will take a completely different form. We firmly believe in the importance of understanding and discriminating which differences have an actual business utility, from those that are due to habit. Regular pruning of obsolete or conflicting processes or terminology is fundamental. However, at the same time, the value of the cross-functional and cross-departmental differences is often overlooked. Rather than imposing a standard vocabulary across the business, we believe that a focus on mapping across glossaries and processes would be the path of least resistance and would enable better collaboration, improving, for example, the time-to-product KPI. This would need a rethink of how ontology alignment tools are enhanced, and moving beyond the existing capabilities to cover cross-functional analogies.

%% file: ejr-onto.tex
\section{Technical enablers}
\label{sec:onto}

This section highlights existing research trends aligned with our vision.

The notion of an ontology for processes and workflows (Section \ref{sec:ontoprocess}) is not new; a prominent example is the formalisation of the Business Process Modelling Notation (BPMN) as an OWL ontology \cite{BPMN-DL-2021}. Nevertheless, the need for suitable infrastructure to facilitate the semantic lift of the already available business processes and workflows is still a limitation.
Approaches like those presented by Abreu et al.~\cite{ontology-workflow-2025} go in the right direction, but may not scale well in dynamic environments. The inclusion of GenAI within the semantic lifting process brings key opportunities to facilitate the adoption and the creation of knowledge graphs for processes and workflows at scale (\eg \cite{DBLP:conf/semweb/GiglouDA23,DBLP:journals/corr/abs-2410-23584}).

Scalability and adaptability depend on moving away from monolithic ontology and knowledge graph solutions, and enabling a modular approach (Section \ref{sec:modular}) with a network of knowledge graphs tailored to the needs of different stakeholders within an organisation. Different business units may have different and possibly inconsistent needs and points of view, and knowledge graphs reflecting these disagreements shall coexist within the network. This heterogeneity may be reflected in the use of different vocabulary, semantic modelling, or following alternative processes.

Advanced knowledge alignment systems (Section \ref{sec:align}) need to move beyond the traditional matching techniques \cite{DBLP:books/daglib/0018324} and enable the discovery and definition of complex semantic correspondences and transformations that bridge knowledge representations across different functional areas of an organisation. Moreover, the alignment process should also be aware of the potential (logical) disagreements and accommodate them to enable collaboration across heterogeneous knowledge graphs from multiple functional units.
For example, in an agentic architecture, multiple agents (\eg one per functional unit) may negotiate semantic correspondences to align their knowledge graphs and reach an agreement that maximises the overall benefit of a cooperative task~\cite{DBLP:conf/kr/Jimenez-RuizPST16,DBLP:journals/corr/abs-2503-15515}.

The ultimate goal is to establish a modular ecosystem of interconnected knowledge graphs, where semantic interoperability can be achieved on demand to support shared understanding and information exchange across diverse business units.